\documentclass[10pt,twocolumn,letterpaper]{article}

\usepackage{wacv}              

\usepackage{multirow}

\definecolor{wacvblue}{rgb}{0.21,0.49,0.74}
\usepackage[pagebackref,breaklinks,colorlinks,allcolors=wacvblue]{hyperref}

\def\wacvPaperID{5} 
\def\confName{WACV}
\def\confYear{2026}

\title{
    SAS-VPReID: A Scale-Adaptive Framework with Shape Priors for Video-based Person Re-Identification at Extreme Far Distances
}

\author{Qiwei Yang, Pingping Zhang\thanks{Corresponding author}, Yuhao Wang, Zijing Gong\\
School of Future Technology, Dalian University of Technology\\
{\tt\small dutyqw@mail.dlut.edu.cn; zhpp@dlut.edu.cn; \{924973292,20221071269\}@mail.dlut.edu.cn}
}

\begin{document}
\maketitle
\begin{abstract}
Video-based Person Re-IDentification (VPReID) aims to retrieve the same person from videos captured by non-overlapping cameras.
At extreme far distances, VPReID is highly challenging due to severe resolution degradation, drastic viewpoint variation and inevitable appearance noise.
To address these issues, we propose a Scale-Adaptive framework with Shape Priors for VPReID, named SAS-VPReID.
The framework is built upon three complementary modules.
First, we deploy a Memory-Enhanced Visual Backbone (MEVB) to extract discriminative feature representations, which leverages the CLIP vision encoder and multi-proxy memory.
Second, we propose a Multi-Granularity Temporal Modeling (MGTM) to construct sequences at multiple temporal granularities and adaptively emphasize motion cues across scales.
Third, we incorporate Prior-Regularized Shape Dynamics (PRSD) to capture body structure dynamics.
With these modules, our framework can obtain more discriminative feature representations.
Experiments on the VReID-XFD benchmark demonstrate the effectiveness of each module and our final framework ranks the first on the VReID-XFD challenge leaderboard.
The source code is available at \url{https://github.com/YangQiWei3/SAS-VPReID}.
\end{abstract}
\begin{figure}
    \centering
    \includegraphics[width=1.0\linewidth]{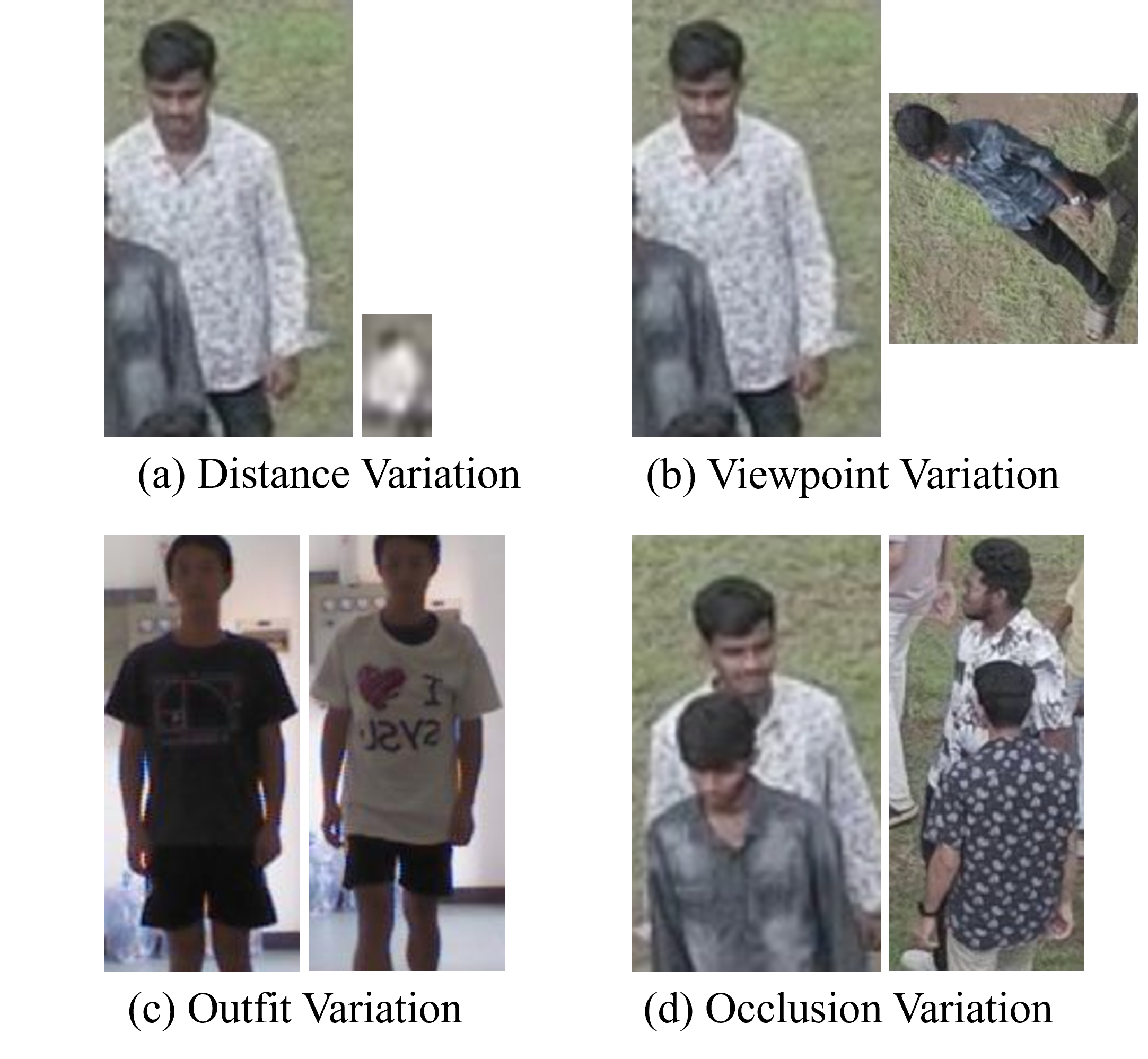}
    \caption{Examples illustrating four major variations in the VReID-XFD challenge: (a) distance variation (different distances leading to noticeable size and clarity differences); (b) viewpoint variation (frontal, side, top-down and tilted views); (c) outfit variation (outfit changes across sessions causing large appearance drifts) and (d) occlusion variation (occlusions and similar-texture backgrounds increasing confusion).}
    \label{fig:dataset}
\end{figure}
\section{Introduction}
\label{sec:introduction}
Person Re-IDentification (ReID) aims to retrieve the same person across non-overlapping cameras.
With great application advantages, it has become a core capability in large-scale intelligence surveillance.
Compared with image-based ReID~\cite{Hambarde2024Covariates,Wang_2025_CVPR,Wang2025DeMo}, video-based ReID~\cite{yu2025climbreid,Yu2025XReID,rashidunnabi2025seeing} exploits richer temporal cues, such as dynamic gait, motion pattern and multi-frame body structure.
However, video-based ReID also introduces additional challenges including temporal redundancy, motion blur, occlusion and complex background dynamics.
Its reliability heavily depends on the quality and stability of observed appearance cues.
Recent progress~\cite{Liu2024DCCT,Liu2023LSTRL,Wang2021PSTA,Liu2021GRL,dai2018video} has largely been driven by stronger spatiotemporal modeling and feature aggregation.

A key assumption of conventional ReID, namely that the same person wears the same clothes, is often violated.
Long-term ReID often involves subjects re-appearing after extended time gaps with substantial outfit changes.
It leads to severe appearance drift and undermines the reliability of color- and texture-dominated features.
This fact motivates the cloth-changing ReID~\cite{qian2020ltcc,wan2020changingclothes,gu2022clothes,li2021learning,loper2015smpl}, which seeks cloth-invariant cues beyond surface appearances and highlights more stable soft-biometric or structural information.

Meanwhile, real-world applications demand cross-platform and cross-view matching, resulting in aerial-ground ReID.
This new task is enabled by UAV and CCTV deployment, where persons captured from drones must be matched to ground cameras for wide-area search and emergency response~\cite{nguyen2023aerial,nguyen2025ag}.
It is inherently difficult due to the extreme viewpoint disparity, resolution mismatch and substantial domain gaps.
Critically, these challenges often occur simultaneously in real-world.
Besides, long-term aerial-ground videos may involve extreme distances and low resolution, along with large viewpoint changes, clothing changes and frequent occlusions, as illustrated in Fig.~\ref{fig:dataset}.
Despite active research, most prior efforts investigate video-based ReID, cloth-changing ReID and aerial-ground ReID in isolation, leaving a gap for the unified setting.
The Video-based Human Recognition at Extreme Far Distances workshop challenge (VReID-XFD) is designed to benchmark exactly this scenario.
It requires tracklet-level retrieval under extreme resolution variations and large cross-domain shifts, where query and gallery tracklets may come from different platforms and different sessions with outfit changes.

A practical solution for VReID-XFD should address four challenges simultaneously: learning strong spatiotemporal feature, keeping robust to appearance drifts and domain gaps, preserving identity-consistent cues and maintaining discrimination under severe low-resolution observations.
Recently, large-scale pre-trained vision-language models like CLIP~\cite{radford2021clip} have shown strong transferability.
However, appearance cues become fragile when the quality of person observations degrades to extreme far distances and clothing changes occur across sessions.
Thus, person appearance features should go beyond the clothing appearance and explicitly incorporate more stable human-centric structural information.

To address these issues, we propose \textbf{SAS-VPReID}, a scale-adaptive framework with shape priors for VPReID.
It jointly strengthens person appearance features, captures multi-granularity temporal dynamics and injects clothing-invariant human-centric cues.
The framework integrates three complementary modules.
First, we deploy a \textbf{Memory-Enhanced Visual Backbone (MEVB)}, which uses video-consistent data augmentation and memory-enhanced supervision to stabilize the model adaptation under extreme far-distance aerial-ground degradation.
Then, we propose a \textbf{Multi-Granularity Temporal Modeling (MGTM)} to efficiently capture multi-granularity  spatiotemporal context by using sequence modeling and scale fusion.
Finally, we incorporate the \textbf{Prior-Regularized Shape Dynamics (PRSD)} to learn cloth-robust identity representations, which leverages SMPL~\cite{loper2015smpl}-based 3D shape cues with an explicit shape prior.
Experiments on the DetReIDXV1 benchmark demonstrate the effectiveness of each module and our final framework ranks the first on the VReID-XFD challenge leaderboard.

In summary, our contributions are as follows:
\begin{itemize}
\item We present SAS-VPReID, a new framework for video-based ReID. It handles the extreme far-distance and cloth-changing challenges in the VReID-XFD setting.
\item We propose MEVB to better adapt CLIP-based representations to the extreme far-distance degradation and clothing change ReID, via video-consistent data augmentation and memory-enhanced supervision.
\item We propose MGTM with sequence modeling and scale fusion to capture multi-granularity spatiotemporal cues from tracklets efficiently.
\item We propose PRSD with 3D shape cues to complement appearance features and improve clothing robustness.
\item Extensive experiments on the DetReIDXV1 benchmark demonstrate the effectiveness of our proposed framework. Meanwhile, our method ranks the first on the VReID-XFD challenge leaderboard.
\end{itemize}
\section{Related Work}
\label{sec:related_work}
\subsection{Video-based Person Re-Identification}
Video-based person ReID aims to match pedestrian tracklets across cameras by jointly exploiting appearance cues and temporal dynamics.
Technically, Wang et al.~\cite{wang2014person} first formulate video-based ReID as a ranking problem to learn discriminative sequence representations.
Zheng et al.~\cite{zheng2016mars} further introduce a large-scale benchmark named MARS that enables deep model training and evaluation.
To better capture temporal information, Fu et al.~\cite{fu2019sta} propose a spatial-temporal attention mechanism to emphasize informative regions and frames.
Dai et al.~\cite{dai2018video} propose the temporal residual learning for feature aggregation.
Beyond attention-based aggregation, Li et al.~\cite{li2019multi} propose to efficiently capture motion patterns with multi-scale temporal cues.
Recently, Wu et al.~\cite{wu2022cavit} propose to perform contextual alignment for long-range temporal modeling, demonstrating the effectiveness of Transformers in video-based ReID.

Recently, large-scale pre-trained vision-language models, e.g., CLIP~\cite{radford2021clip}, have been introduced to improve the generalization in ReID.
For example, Li et al.~\cite{li2023clip} propose to transfer CLIP knowledge to ReID in a parameter-efficient manner.
Yu et al.~\cite{yu2025climbreid} further enhance video-based ReID by combining CLIP representations with efficient sequence modeling.
Although these methods~\cite{Liu2023LSTRL,Liu2024TMT,Liu2024DCCT,Yu2024TFCLIP,Yu2025XReID} achieve remarkable performance on conventional settings, they often assume relatively high imaging quality and may struggle with the severe, instance-dependent degradations in extremely long-range videos.
In contrast, our SAS-VPReID is tailored for such challenging scenarios by jointly strengthening transferable appearance learning, multi-granularity temporal modeling and cloth-invariant structural dynamics.
\subsection{Cloth-changing Person Re-Identification}
Cloth-changing person ReID aims to match the same person across time when outfits vary, which causes severe appearance drifts and weakens color cues.
To support this setting, Qian et al.~\cite{qian2020ltcc} introduce the long-term cloth-changing benchmark named LTCC.
Wan et al.~\cite{wan2020changingclothes} collect VC-Clothes to emphasize clothing variations.
Gu et al.~\cite{gu2022clothes} further extend the challenge to videos.
To mitigate clothing-induced bias, existing methods mainly explore three directions.
First, many methods~\cite{Yu2025HPL} learn clothing-invariant representations by disentangling clothing-related cues from person features.
Second, some methods~\cite{li2021learning} exploit soft-biometric cues so that identity matching relies less on clothing appearance.
Third, some methods introduce explicit geometry priors.
For example, Nguyen et al.~\cite{nguyen2024temporal} leverage SMPL-related parameters as biometric evidence under clothing changes.
Although existing methods show promising results, they often rely on high-quality observations with sufficient resolution and limited viewpoint variations.
Thus, they may fail when soft-biometric cues are corrupted by extreme blur and low resolution.
In contrast, we address far-distance aerial-ground and cloth-changing ReID by enhancing appearance features and incorporating SMPL-based structural cues for clothing-invariant representations.
\subsection{Aerial-Ground Person Re-Identification}
Aerial-ground person ReID aims to retrieve persons across aerial and ground cameras, facing drastic viewpoint changes, resolution mismatches and large domain gaps.
Recent works~\cite{nguyen2023aerial,nguyen2025ag} establish benchmarks and baselines for this task.
Beyond dedicated benchmarks, cross-platform adaptation \cite{Zhang2024CrossPlatform} and multi-temporal cross-view learning \cite{rashidunnabi2025seeing} further promote discriminative representations.
However, most approaches~\cite{Hu2025LATex,Hu2025SDReID} still rely heavily on appearance cues, which deteriorate under extreme far-distance degradation and outfit variations.
In contrast, our SAS-VPReID strengthens discriminative appearances with CLIP and complements it with shape dynamics for robust retrieval.
\begin{figure*}
\centering
\includegraphics[width=1.0\textwidth]{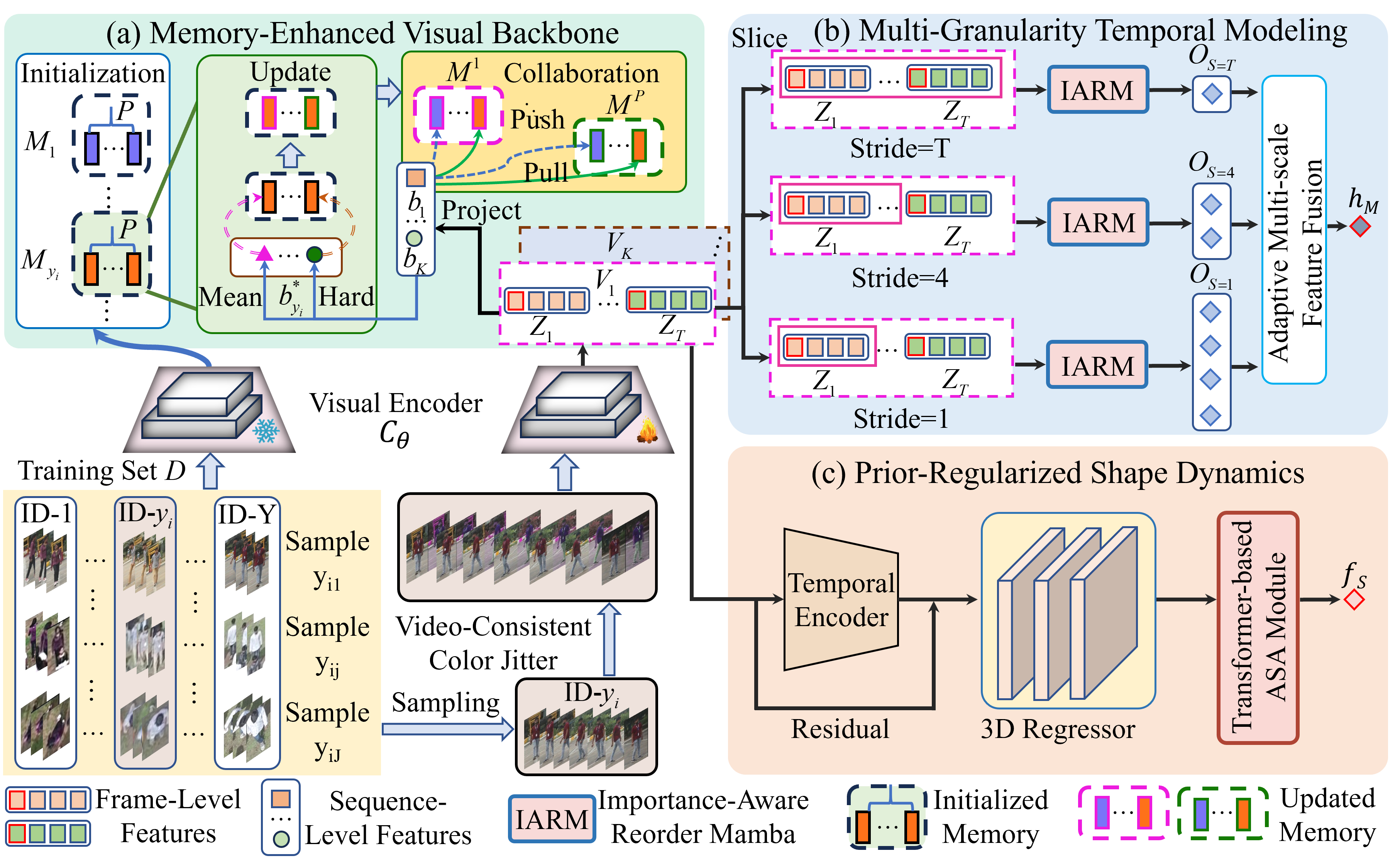}
\caption{Illustration of our proposed SAS-VPReID framework for the VReID-XFD Challenge. The Memory-Enhanced Visual Backbone (MEVB) strengthens the visual feature representation, with video-consistent data augmentation and memory-based supervision to stabilize the model adaptation under extreme far-distance degradation. Then, the Multi-Granularity Temporal Modeling (MGTM) captures multi-granularity spatiotemporal features via efficient sequence modeling and learnable scale fusion. Finally, the Prior-Regularized Shape Dynamic (PRSD) injects clothing-invariant identity cues by modeling SMPL-based shape parameters over time, with an explicit shape prior and temporal aggregation. With the integration of MEVB, MGTM and PRSD, our SAS-VPReID effectively learns discriminative multi-granularity spatiotemporal and clothing-invariant features for extreme far-distance video-based person ReID.}
\label{fig:overall}
\end{figure*}
\section{Methodology}
\label{sec:method}
In this work, we present a solution for the VReID-XFD Challenge, where video tracklets are captured at extreme far distances with severe resolution degradation, obvious aerial-ground domain gaps and cross-session clothing changes.
In this competition setting, frame-level appearance cues become highly unstable and often insufficient for reliable matching.
Therefore, we advocate a unified design that can retain discriminative identity evidence, capture multi-granularity temporal dynamics and incorporate clothing-invariant cues to improve robustness in such challenging scenarios.
As illustrated in Fig.~\ref{fig:overall}, we propose SAS-VPReID, a scale-adaptive framework with shape priors.
It consists of three complementary modules, i.e., Memory-Enhanced Visual Backbone (MEVB), Multi-Granularity Temporal Modeling (MGTM) and Prior-Regularized Shape Dynamics (PRSD).
These modules jointly produce a robust feature descriptor for extreme far-distance aerial-ground and cloth-changing ReID.
The detailed designs are elaborated in the following subsections.
\subsection{Task Definition}
\label{sec:task_definition}
Given a video tracklet $V_i=\{F_t\}_{t=1}^{T}$ of length $T$, with the identity label $y_i\in\{1,\dots,Y\}$, the goal of video-based person ReID is to learn an embedding $e_i\in\mathbb{R}^{d}$ such that samples of the same identity are close while those of different identities are well separated.
In the VReID-XFD challenge, the query and gallery tracklets may come from different platforms (aerial vs.\ ground) and different sessions with clothing changes.
It requires the learned representation to remain discriminative and robust.
\subsection{Memory-Enhanced Visual Backbone}
\label{sec:mmc}
Extreme far-distance aerial-ground videos substantially undermine the reliability of appearance cues, since persons often occupy only a few pixels.
The cross-platform imaging gap and cross-session clothing changes further amplifies intra-class variations.
Under such conditions, a strong visual backbone should provide discriminative feature representations to  simultaneously retain identity evidence and support stable optimization.

To this end, we enhance the visual backbone from three complementary aspects, namely backbone capacity upgrade, video-consistent data augmentation and memory-enhanced supervision.
They form a memory-enhanced visual backbone that provides more discriminative and robust video representations for subsequent modules.

\textbf{Backbone Capacity Upgrade.}
We employ the pre-trained CLIP visual encoder~\cite{radford2021clip} as the backbone $C_{\theta}$ due to its strong generalization.
To better preserve discriminative cues in extremely low-resolution observations, we use ViT-L to increase the representation capacity.
Given an input frame $F_t$, the visual backbone outputs token features:
\begin{equation}
g_t = C_{\theta}(F_t) = [g^{cls}_t, g^1_t, \dots, g^N_t],
\label{eq:mem_backbone_tokens}
\end{equation}
where $g^{cls}_t$ is the class token and $\{g^n_t\}_{n=1}^{N}$ are patch tokens.
Then, a linear projection is used to map $g^{cls}_t$ to the unified embedding space, yielding frame-level features $f_t \in \mathbb{R}^d$.
$d$ denotes the feature dimension.

\textbf{Video-Consistent Data Augmentation.}
To increase the appearance diversity without breaking video continuity, we adopt a video-consistent color jitter operation $\mathcal{A}$ during training.
For each sampled tracklet, we randomly draw one color jitter parameter as a hue-shift factor $\phi \sim \mathcal{U}(-h, h)$.
With the probability $p_c$, we apply it consistently to all frames in the tracklet; otherwise we keep the original frames unchanged.
The process can be present as follows:
\begin{equation}
\mathcal{A}(F_t,\phi) = adjust\_hue(F_t,\phi)
\label{eq:vc_cj_A}
\end{equation}
\begin{equation}
\tilde{F}_t =
\begin{cases}
\mathcal{A}(F_t,\phi), & \text{with probability } p_c, \\
        F_t,                   & \text{otherwise}.
\end{cases}
\label{eq:vc_cj}
\end{equation}
This video-consistent design preserves temporal coherence and prevents the model from learning spurious frame-to-frame color fluctuations, which is critical for video-based ReID under clothing, illumination and style changes.

\textbf{Memory-Enhanced Supervision.}
Even with a strong backbone, directly fine-tuning on extreme far-distance videos can be unstable, since the same identity may present different viewpoints, blur levels and clothing states.
Therefore, we introduce a lightweight memory mechanism~\cite{yu2025climbreid} to provide identity-consistent anchors for the backbone optimization.
Specifically, we first form a sequence-level embedding by the temporal average pooling:
\begin{equation}
v_i = \frac{1}{T}\sum_{t=1}^{T} f_t.
\label{eq:seq_avg}
\end{equation}
For each identity $y$, we maintain a multi-proxy memory set $\{M^p_y\}_{p=1}^{P}$, where $P$ denotes the number of proxies, and each proxy captures a distinct mode.
During training, the proxies are updated online with a momentum $\mu$:
\begin{equation}
M^p_y \leftarrow \mu M^p_y + (1-\mu)\, v^{*}_y,
\label{eq:mem_update}
\end{equation}
where $\mu \in [0,1)$ is the momentum coefficient and $v^{*}_y$ is a representative sequence feature selected from the current mini-batch of identity $y$.
To encourage the proxy diversity and improve the coverage of intra-class variations, we update proxies using complementary representatives, i.e., a mean feature capturing the dominant mode and a hard feature capturing challenging variations.

Given a sample feature $v_i$ with label $y_i$, we optimize the backbone with a memory-based contrastive objective:
\begin{equation}
\mathcal{L}^{(i)}_{\text{me}}= -\log
    \frac{\sum_{p=1}^{P} \exp\left(\langle v_i, M^p_{y_i}\rangle/\tau\right)}
    {\sum_{y=1}^{Y}\sum_{p=1}^{P}\exp\left(\langle v_i, M^p_{y}\rangle/\tau\right)},
\label{eq:mem_nce}
\end{equation}
where $\tau$ is the temperature parameter and $\langle\cdot,\cdot\rangle$ denotes the cosine similarity after $\ell_2$ normalization.
This objective enforces that each sample feature aligns with a set of proxies rather than a single prototype.
It is particularly beneficial when the same person exhibits large mode shifts across views, distances and sessions.
Importantly, the memory introduces no heavy architectural overhead.
It acts as an external, online-updated supervisory signal, which enhances the backbone's discrimination and training stability under extreme degradations.

In summary, by combining a high-capacity visual encoder, video-consistent data augmentation and multi-proxy memory-enhanced supervision, our MEVB yields stronger and more robust video representations, serving as a reliable foundation for subsequent modules.
\subsection{Multi-Granularity Temporal Modeling}
In the VReID-XFD challenge, representing extreme far-distance videos require temporal modeling because single-frame appearance is often degraded by low resolution, blur and clutter.
Moreover, large platform motions can make tracklets unstable, causing discriminative cues to appear only intermittently.
Therefore, instead of relying on a single temporal receptive field, we aim to learn a representation that can capture both fine-grained short-term variations like immediate pose transitions and coarse-grained long-term consistency like stable motion tendency.
Inspired by the previous work~\cite{yu2025climbreid}, we achieve this via a multi-granularity temporal modeling module, which extracts temporal features at multiple time scales and fuses them adaptively for robust video-level representations.

Given a tracklet, our MEVB can produce frame-level features $\{f_t\}_{t=1}^{T}$.
Our goal is to convert this sequence into a discriminative video representation.

\textbf{Multi-Granularity Temporal Slicing.}
To capture multi-granularity temporal information, we construct multiple sub-sequences with different sampling strides.
For each stride $s \in \mathcal{S}$, we build a down-sampled sequence:
\begin{equation}
\mathbf{F}^{(s)} = \{[f_1, f_2, \dots,f_{s}],\dots,[f_{T-s}, \dots,f_{T}]\},
\label{eq:multi_stride_seq}
\end{equation}
where a small $s$ retains fine temporal details and a larger $s$ provides a longer temporal context.
This design is especially beneficial in video-based ReID, where frame-level evidence may be weak and multi-granularity temporal aggregation helps extract robust identity features.

\textbf{Efficient Temporal Modeling.}
For each temporal granularity, we apply an efficient sequence model to capture temporal dependencies.
Concretely, we employ a Mamba-based temporal operator due to its linear complexity with respect to the sequence length~\cite{gu2024mamba}.
It is beneficial for processing large-scale tracklets.
To further prioritize informative temporal content, we reorder tokens based on their relevances to the class token and process the sequence in different directions to strengthen context aggregation:
\begin{equation}
\begin{aligned}
\mathbf{H}^{(s)} =& \mathrm{Mamba}\big(\mathrm{Reorder}(\mathbf{F}^{(s)})\big)\\
                  & + \mathrm{Mamba}\big(\mathrm{Reorder}(\mathbf{F}^{(s)})^{\mathrm{rev}}\big).
\end{aligned}
\label{eq:mamba_bidir}
\end{equation}
Then, the temporal average pooling operator aggregates $\mathbf{H}^{(s)}$ into a stride-specific video feature $h^{(s)} \in \mathbb{R}^d$.

As observed, this multi-granularity design realizes that fine-grained streams capture short-lived cues that may appear immediately, while coarse-grained streams enforce global consistency and reduce sensitivity to frame noises.

\textbf{Learnable Multi-Granularity Fusion.}
In the previous work~\cite{yu2025climbreid}, multi-stride features are combined by a simple average, which implicitly assumes all temporal strides are equally informative for every tracklet.
However, this assumption is fragile in VReID-XFD.
The optimal temporal stride depends on the tracklet quality, motion speed and usable identity evidence.
Therefore, we introduce a learnable scale fusion mechanism that allows the model to adaptively select useful temporal granularity.
The process can be expressed as follows:
\begin{equation}
h_{\text{M}} = \sum_{s \in \mathcal{S}} w_s \, h^{(s)},
\label{eq:learnable_scale_fusion}
\end{equation}
\begin{equation}
w_s = \frac{\exp(a_s)}{\sum_{k \in \mathcal{S}} \exp(a_k)},
\label{eq:learnable_scale_fusion}
\end{equation}
where $\{a_s\}$ are learnable parameters.
This adaptive fusion is especially advantageous, as it mitigates sensitivity to temporal jitter by dynamically prioritizing the most reliable temporal stride.
Moreover, it enhances feature discrimination by leveraging long-range contextual cues, while preserving informative short-term motion patterns.

Overall, the proposed MGTM produces a robust temporal feature $h_M$ that complements the appearance feature from the visual backbone and serves as a reliable supplement for subsequent 3D-aware shape modeling.
\subsection{Prior-Regularized Shape Dynamics}
\label{sec:prior_shape_dynamics}
In the VReID-XFD challenge, extreme far-distance cloth-changing tracklets jointly exhibit drastic viewpoint distortions and cross-session clothing changes.
In such a challenging setting, a reliable feature representation should go beyond surface appearances and incorporate human-centric structural information that remain comparatively stable across outfits and style domains.
Motivated by this requirement, we introduce Prior-Regularized Shape Dynamics that inject compact 3D body-shape cues into our framework.
Following~\cite{nguyen2024temporal}, rather than estimating detailed poses and meshes that are hard to recover from low-quality observations, we regress SMPL~\cite{loper2015smpl}-based shape parameters and explicitly model their temporal dynamics.
As a result, it provides a clothing-invariant complement to typical appearance-driven features.

\textbf{Per-Frame Shape Parameterization.}
Given the frame-level features $\{f_t\}_{t=1}^{T}$, we first inject temporal contexts into each feature using a lightweight temporal encoder $G(\cdot)$.
It is implemented by GRU~\cite{cho-etal-2014-learning} layers and preserves the original information via a residual connection:
\begin{equation}
f_t^g = G(f_t),
\end{equation}
\begin{equation}
\hat{f}_t^g = f_t + f_t^g.
\end{equation}
Then, a compact regressor $R(\cdot)$ is used to predict the SMPL-based shape parameters:
\begin{equation}
\alpha_t = R(\hat{f}_t^g),
\end{equation}
where $\alpha_t \in \mathbb{R}^{10}$ encodes identity-related body proportions that are less sensitive to clothing variations.

\textbf{Shape Dynamics.}
Per-frame shape predictions can be unstable due to occlusion and motion blur.
Thus, directly averaging $\{\alpha_t\}$ may over-trust noisy frames.
To address this issue, we model shape dynamics in a two-stage manner.
More specifically, we first apply a GRU-based smoothing module $G'$ to suppress frame-wise jitters.
Then, we use a Transformer-based temporal modeling module to capture long-range dependencies, which is particularly important when informative frames appear intermittently.
Concretely, we obtain temporally smoothed features as follows:
\begin{equation}
\tilde{\alpha}_t = G'(\alpha_t, \tilde{\alpha}_{t-1}).
\end{equation}
Then, we feed the sequence into a 4-layer Transformer encoder to perform global temporal interactions.
\begin{equation}
\{\bar{\alpha}_t\}_{t=1}^{T} = \mathrm{TransEnc}\big(\{\tilde{\alpha}_t\}_{t=1}^{T}\big).
\end{equation}
Finally, we perform the temporal average pooling to form a sequence-level shape representation:
\begin{equation}
f_S = \frac{1}{T}\sum_{t=1}^{T}\bar{\alpha}_t.
\end{equation}

\textbf{Prior-Regularized Optimization.}
In cloth-changing videos, a key challenge is that regressing 3D-related shape parameters can drift without strong constraints.
Instead of using batch statistics as an implicit regularizer, we impose an explicit SMPL shape prior to encourage physically plausible solutions and stabilize training.
Let $\alpha_{\mathrm{SMPL}}\in\mathbb{R}^{10}$ denote the canonical mean-shape parameters from SMPL. We apply an $\ell_2$ prior constraint:
\begin{equation}
    \mathcal{L}_{\alpha}=\frac{1}{T}\sum_{t=1}^{T}\lVert \bar{\alpha}_t-\alpha_{\mathrm{SMPL}}\rVert_2^2 .
\end{equation}
This prior-regularized formulation is particularly beneficial when observations are noisy and under-determined.
\subsection{Training and Inference}
\label{sec:train_infer}
\textbf{Loss Functions.}
We impose widely-used discriminative loss functions on the temporal representation $h_M$ and the shape-aware representation $f_S$.
Specifically, we adopt the label-smoothed classification loss and the triplet loss:
\begin{equation}
\mathcal{L}_{id}=-\log p(y_i \mid e_i),
\label{eq:id_loss}
\end{equation}
\begin{equation}
\mathcal{L}_{tri}=\big[m + d(e_i,e_i^{+}) - d(e_i,e_i^{-})\big]_{+},
\label{eq:tri_loss}
\end{equation}
where $e_i$ denotes the corresponding representation, $d(\cdot,\cdot)$ is a distance metric and $m$ is a margin.
The full training loss is expressed as follows:
\begin{equation}
    \mathcal{L}=\mathcal{L}_{tri}
    +\lambda_{id}\mathcal{L}_{id}
    +\lambda_{me}\mathcal{L}_{me}
    +\lambda_{\alpha}\mathcal{L}_{\alpha},
    \label{eq:total_loss}
\end{equation}
where $\lambda_{id}$, $\lambda_{me}$ and $\lambda_{\alpha}$ are balancing coefficients.

\textbf{Optimization.}
To stabilize the model training in the aerial-ground domain gap and severe resolution degradation in VReID-XFD, we apply differential learning rates, namely a smaller rate for the pre-trained CLIP backbone and larger rates for newly introduced modules, such as MGTM, PRSD and projection heads.

\textbf{Inference.}
At test time, we use a fused tracklet descriptor for retrieval.
Concretely, we concatenate the clip-level appearance feature $v_i$, the multi-granularity temporal feature $h_M$ and the shape-aware feature $f_S$.
\begin{table*}[t]
    \centering
    \caption{Performance comparison on DetReIDXV1 benchmark. We report mAP and CMC (Rank-1/5/10) under three evaluation settings. Best results in each metric are highlighted in bold.}
    \label{tab:sota_video}
    \resizebox{\textwidth}{!}{
        \begin{tabular}{l|cccc|cccc|cccc}
            \hline
            \multirow{2}{*}{Method}              &
            \multicolumn{4}{c|}{A$\rightarrow$G} &
            \multicolumn{4}{c|}{G$\rightarrow$A} &
            \multicolumn{4}{c}{A$\rightarrow$A}                                                                                                                                                                                                              \\
            \cline{2-5}\cline{6-9}\cline{10-13}
                                                 & mAP            & R1             & R5             & R10            & mAP            & R1             & R5             & R10            & mAP            & R1             & R5             & R10            \\
            \hline
            PSTA~\cite{Wang2021PyramidSTA}       & 34.40          & 22.30          & 46.90          & 59.70          & 17.00          & 40.40          & 56.20          & 59.60          & 10.50          & 13.00          & 23.80          & 30.30          \\
            BiCNet-TKS~\cite{Hou2021BicnetTKS}   & 33.28          & 21.71          & 44.85          & 59.30          & 22.12          & 41.57          & 58.43          & 65.17          & 9.71           & 13.30          & 26.78          & 36.57          \\
            SINet~\cite{Bai2022SalientToBroad}   & 38.46          & 25.62          & 52.47          & 66.57          & 16.98          & 23.50          & 49.44          & 59.55          & 12.85          & 14.06          & 24.51          & 30.94          \\
            VSLA~\cite{Zhang2024CrossPlatform}   & 41.63          & 28.96          & 54.71          & 69.13          & 26.26          & 58.43          & 65.17          & 69.66          & 13.83          & 15.96          & 26.10          & 32.77          \\
            \hline
            Yu Fan Lin                           & 29.17          & 22.85          & 48.16          & 62.80          & 14.47          & 26.97          & 44.94          & 52.81          & 9.80           & 11.04          & 20.44          & 27.00          \\
            Rajbhandari Ashwat ASU               & 38.76          & 32.37          & 59.43          & 74.75          & 28.36          & 62.92          & 75.28          & 77.53          & 16.02          & 18.68          & 29.97          & 38.29          \\
            JNNCE ISE                            & 38.50          & 32.05          & 59.21          & 73.06          & 29.48          & 64.04          & 68.54          & 70.79          & 16.52          & 19.00          & 30.05          & 36.82          \\
            CJKang                               & 39.63          & 33.65          & 57.68          & 73.38          & 28.59          & 62.92          & 69.66          & 75.28          & 16.81          & 20.08          & 31.19          & 38.54          \\
            PPCUITK                              & 39.21          & 32.77          & 59.64          & 71.97          & 29.75          & 62.92          & 69.66          & 73.03          & 16.66          & 19.15          & 30.30          & 36.86          \\
            H Nguyn\_UIT                         & 39.59          & 33.15          & 60.49          & 74.88          & 34.49          & 62.92          & 66.29          & 69.66          & 19.79          & 20.37          & 29.08          & 34.85          \\
            \hline
            \textbf{SAS-VPReID(Ours)}            & \textbf{43.93} & \textbf{37.77} & \textbf{65.26} & \textbf{75.31} & \textbf{35.44} & \textbf{69.66} & \textbf{80.90} & \textbf{87.64} & \textbf{20.13} & \textbf{25.39} & \textbf{39.58} & \textbf{48.44} \\
            \hline
        \end{tabular}
    }
\end{table*}
\section{Experimental Setups}
\subsection{Datasets and Evaluation Protocols}
\textbf{Datasets.}
We conduct extensive experiments on DetReIDXV1, a video-based person ReID benchmark derived from the DetReIDX~\cite{hambarde2025detreidx}.
Specifically, DetReIDX is designed to expose failure modes under extreme capture conditions, where person ROIs can degrade to sub-10-pixel silhouettes with drastic viewpoint and appearance variations.
It contains 509 identities and over 13M manually annotated bounding boxes.
Each identity is recorded in two sessions collected on different days with clothing and illumination changes.
For the aerial part, UAV videos are captured with altitude up to 120 m and distance 10-120 m, covering 18 UAV viewpoints per session and three pitch angles (30°/60°/90°), yielding severe scale variation and top-view distortion.
We follow the official DetReIDXV1 setting, which provides extreme far-distance aerial-ground ReID data across two sessions with varying outfits.

\textbf{Evaluation Protocols.}
Following common practices in video-based person ReID, we report the Cumulative Matching Characteristic (CMC) at Rank-K and the mean Average Precision (mAP).
The video-based person ReID task is formulated as tracklet-level retrieval, namely ranking gallery tracklets for each query tracklet by the similarity.
The DetReIDXV1 benchmark defines three independent evaluation settings, to assess cross-platform and intra-aerial robustness under extreme far-distance degradations, including Aerial-to-Ground (A$\to$G), Ground-to-Aerial (G$\to$A) and Aerial-to-Aerial (A$\to$A).
The first two settings measure cross-platform retrieval across aerial and ground cameras with substantial viewpoint and imaging-domain gaps, while the last setting focuses on the robustness within the aerial domain under varying viewpoints and sessions.
For each setting, we compute CMC, mAP and mAP-3 based on the ranked retrieval results over all queries.
We report the corresponding performance to summarize the tracklet-level matching accuracy and overall ranking quality.
Here, mAP-3 denotes the average mAP over the three evaluation settings (A$\rightarrow$G, G$\rightarrow$A, A$\rightarrow$A).
\subsection{Implementation Details}
The proposed framework is implemented by using PyTorch.
We adopt CLIP-ViT-L/14~\cite{radford2021clip} as the visual encoder for stronger representations.
We apply random flipping and random erasing, and introduce video-consistent color jitter by sharing the same jitter parameters across all frames in a sampled tracklet.
Each frame is resized to $252 \times 126$.
We train for 40 epochs with Adam~\cite{adam2014method}.
The batch size is 16, including 4 identities and 4 tracklets per identity, with 8 frames per tracklet.
The learning rate is set to $3.5 \times 10^{-4}$ for randomly initialized modules and $5 \times 10^{-6}$ for pre-trained parts.
We warm up for 10 iterations from $3.5 \times 10^{-5}$ to $3.5 \times 10^{-4}$ and decay it by 0.1 at epochs 10, 20 and 30.
For MEVB, we use mean and hard samples, maintain $P=2$ proxies per identity with momentum $\mu=0.2$. We set $h = 0.3$ and $p_c = 0.5$.
For MGTM, we fix the slice stride set as $S=[2,4,8]$ and set $\tau=1.0$, $\lambda_{id}=0.25$ and $\lambda_{me}=\lambda_{\alpha}=1.0$.
\begin{table}[t]
    \centering
    \caption{Leaderboard results on DetReIDXV1 benchmark for the VReID-XFD challenge. We report the official mAP-3 scores of the top submissions.}
    \label{tab:leaderboard}
    \begin{tabular}{l r}
        \toprule
        Team                               & mAP-3          \\
        \midrule
        rashid                             & 15.29          \\
        Yu Fan Lin                         & 20.07          \\
        VSLA~\cite{Zhang2024CrossPlatform} & 28.11          \\
        Rajbhandari Ashwat (ASU)           & 28.18          \\
        JNNCE -ISE                         & 28.28          \\
        PPCU-ITK                           & 28.73          \\
        CJKang                             & 29.00          \\
        Ha Nguyen\_UIT                     & 30.43          \\
        \textbf{DUT\_IIAU\_LAB(Ours)}      & \textbf{32.89} \\
        \bottomrule
    \end{tabular}
\end{table}
\section{Experimental Results}
\subsection{Comparison with State-of-the-Arts}
As reported in Tab.~\ref{tab:sota_video} and Tab.~\ref{tab:leaderboard}, our SAS-VPReID achieves the best performance on all three evaluation settings.
It consistently outperforms both representative baselines and the top-ranked VReID-XFD challenge submissions under the extreme far-distance aerial-ground and cloth-changing setting.
On the most challenging cross-platform setting A$\rightarrow$G, our method reaches 43.93 mAP and 37.77 Rank-1, exceeding the strongest baseline VSLA by +2.30 mAP and improving over the best challenge Rank-1 result by +4.12.
On G$\rightarrow$A, we further obtain 35.44 mAP and 69.66 Rank-1, surpassing the best competing mAP by +0.95 and the best Rank-1 by +5.62.
Notably, even on the intra-aerial setting A$\rightarrow$A, our method remains clearly superior, achieving 20.13 mAP and 25.39 Rank-1, which improves upon the best competitor by +0.34 mAP and +5.02 Rank-1.
Overall, our submission (DUT\_IIAU\_LAB) achieves the best performance on the VReID-XFD leaderboard with 32.89 mAP-3, outperforming the runner-up by 2.46 points.

Our best performance comes from a unified design that improves the training stability and temporal consistency in data augmentation under severe degradation and domain gaps.
Meanwhile, it also enhances identity-discriminative cues beyond appearances for cloth-changing scenarios.
\begin{table}[t]
    \centering
    \caption{Module ablation on DetReIDXV1. ``Baseline'' is CLIMB-ReID. We progressively add MDLR, VC-CJ, MGTM and PRSD.}
    \label{tab:ablation_components}
    \resizebox{\linewidth}{!}{
        \begin{tabular}{c|cccc|c}
            \hline
            Model       & MDLR       & VC-CJ      & MGTM       & PRSD       & mAP-3          \\
            \hline
            1(Baseline) &            &            &            &            & 30.79          \\
            2           & \checkmark &            &            &            & 31.77          \\
            3           & \checkmark & \checkmark &            &            & 32.54          \\
            4           & \checkmark & \checkmark & \checkmark &            & 32.78          \\
            5 (Ours)    & \checkmark & \checkmark & \checkmark & \checkmark & \textbf{32.89} \\
            \hline
        \end{tabular}
    }
\end{table}
\subsection{Ablation Study}
\label{sec:ablation}
\subsubsection{Component Analysis}
We perform ablation studies on DetReIDXV1 to quantify the contribution of each module.
As summarized in Tab.~\ref{tab:ablation_components}, we start from the CLIMB-ReID~\cite{yu2025climbreid} as the baseline (Model~1) and progressively add Module-wise Differential Learning Rates (MDLR), Video-Consistent Color Jitter (VC-CJ), Multi-Granularity Temporal Modeling (MGTM) and Prior-Regularized Shape Dynamics (PRSD).

\textbf{Effectiveness of MDLR.}
With MDLR (Model~2), the value of mAP-3 increases from 30.79 to 31.77 (+0.98), indicating that stabilized fine-tuning, namely smaller learning rate for the pre-trained backbone and larger learning rates for new modules, is important under extreme far-distance aerial-ground shifts.

\textbf{Effectiveness of VC-CJ.}
With VC-CJ, our method further boosts mAP-3 from 31.77 to 32.54 (\(+0.77\)).
By sharing identical hue jitter parameters across frames, VC-CJ can preserve temporal coherence, prevent spurious frame-to-frame color shifts, and improves the robustness to cloth-changing, illumination and camera-style variations.

\textbf{Effectiveness of MGTM.}
Incorporating MGTM yields a consistent gain from 32.54 to 32.78 (\(+0.24\)), suggesting that multi-granularity temporal modeling provides complementary refinement beyond optimization and augmentation.

\textbf{Effectiveness of PRSD.}
Finally, adding PRSD further improves mAP-3 from 32.78 to 32.89 (\(+0.11\)).
Although modest, the gain indicates that shape cues with an explicit prior can complement appearance features when clothing changes and severe blur make appearance cues unreliable.
\subsubsection{Comparison of PRSD and TSM.}
To evaluate whether our PRSD is superior to the previous Temporal Shape Modeling (TSM)~\cite{nguyen2024temporal}, we conduct a controlled comparison on DetReIDXV1 by switching the temporal shape branch while keeping other settings unchanged.
As reported in Tab.~\ref{tab:ablation_components}, replacing TSM with PRSD improves mAP-3 from 32.54 to 32.89 (+0.35).
This result indicates that explicitly regularizing shape dynamics with a prior over SMPL shape parameters, together with Transformer-based temporal modeling, can produce more discriminative tracklet representations in extremely far-distance scenarios.
\begin{table}[t]
    \centering
    \caption{Performance comparison of TSM and PRSD on DetReIDXV1 benchmark.}
    \label{tab:ablation_components}

    \begin{tabular}{c|cc|c}
        \hline
        Model & TSM        & PRSD       & mAP-3          \\
        \hline
        1     & \checkmark &            & 32.54          \\
        2     &            & \checkmark & \textbf{32.89} \\
        \hline
    \end{tabular}
\end{table}
\begin{table}[t]
    \centering
    \caption{Computational cost and scalability analysis under different backbone scales (ViT-B vs. ViT-L). We report inference speed, GPU memory allocation, FLOPs and total parameter count using the same framework and settings, with only the backbone replaced.}
    \label{tab:benchmark}
    \begin{tabular}{ccc}
        \hline
        Metric                   & ViT-B  & ViT-L   \\
        \hline
        Inference Speed(ms/iter) & 325.48 & 342.58  \\
        GPU Memory Allocate(M)   & 782.49 & 1694.81 \\
        FLOPs Forward(G)         & 74.26  & 296.76  \\
        Total Params(M)          & 196.74 & 435.90  \\
        \hline
    \end{tabular}
\end{table}
\subsection{Computational Cost and Scalability}
Tab.~\ref{tab:benchmark} shows that scaling the backbone from ViT-B to ViT-L substantially increases the model capacity and computational/memory cost, while causing only a minor increase in inference latency.
This suggests our framework scales well with stronger backbones, making ViT-L a practical choice when prioritizing robustness and accuracy in extreme far-distance ReID, like VReID-XFD challenge.
\section{Conclusion}
In this paper, we propose SAS-VPReID for the VReID-XFD challenge.
Our method addresses extreme far-distance aerial-ground ReID by combining video-consistent color jitter, a CLIP backbone with memory-enhanced supervision, multi-granularity temporal modeling with learnable scale fusion and a SMPL-based prior-regularized temporal 3D shape branch.
Experiments on the DetReIDXV1 benchmark demonstrate the effectiveness of our modules and our final framework ranks the first place on the VReID-XFD challenge leaderboard.
\section*{Acknowledgements}
This work was supported in part by the National Natural Science Foundation of China (No.62576069) and Natural Science Foundation of Liaoning Province (No.2025-MS-025).
We thank all the authors of CLIMB-ReID for their insightful work and making their code publicly available.

{
\small
\bibliographystyle{ieeenat_fullname}
\bibliography{main}
}
\end{document}